\pdfoutput=1
\documentclass[11pt]{article}

\usepackage{ACL2023}

\usepackage{times}
\usepackage{latexsym}

\usepackage[T1]{fontenc}
\usepackage[utf8]{inputenc}

\usepackage{microtype}

\usepackage{inconsolata}

\DeclareCaptionStyle{ruled}{labelfont=normalfont,labelsep=colon,strut=off} % DO NOT CHANGE THIS
\usepackage{helvet}  % DO NOT CHANGE THIS
\usepackage{courier}  % DO NOT CHANGE THIS

\usepackage{booktabs}
\usepackage{mathtools}
\usepackage{varwidth}
\usepackage{empheq}
\usepackage{amssymb}
\usepackage{tabularx}
\usepackage{multirow}
\usepackage{epsfig}
\usepackage{amsmath}
\usepackage{cancel}
\usepackage{amsfonts}       % blackboard math symbols
\usepackage{nicefrac}       % compact symbols for 1/2, etc.
\usepackage{bbm}
\usepackage{bm}
\usepackage{subcaption}
\usepackage{defs}
\usepackage{enumitem}% http://ctan.org/pkg/enumitem

\usepackage{soul}
\usepackage{url}
\usepackage{hyperref}
\usepackage{dirtytalk}
\usepackage{amsfonts}

\usepackage{algorithm}

\usepackage{algpseudocode}
\usepackage{array,multirow}
\usepackage{xcolor}
\usepackage{amsthm}

\usepackage[colorinlistoftodos]{todonotes}
\usepackage{enumitem}
\newlist{Properties}{enumerate}{2}
\setlist[Properties]{label=Property \arabic*.,itemindent=*}

\usepackage{nicefrac}       % compact symbols for 1/2, etc.

\renewcommand{\vec}[1]{\boldsymbol{#1}}
\renewcommand{\vec}[1]{\boldsymbol{#1}}

\usepackage{siunitx} % to round numbers
\renewcommand{\vec}[1]{\boldsymbol{#1}}
\usepackage{algpseudocode}
\usepackage{newfloat}
\usepackage{listings}

\title{Self-Distilled Quantization: Achieving High Compression Rates in Transformer-Based Language Models}

\author{James O' Neill \and Sourav Dutta \\
  Huawei Ireland Research Center \\
  Georges Court, Townsend St, Dublin 2, Ireland \\
  \texttt{james.o.neil@huawei-partners.com}, \\ \texttt{sourav.dutta2@huawei.com}}

\begin{document}
\maketitle

\begin{abstract}
We investigate the effects of post-training quantization and quantization-aware training on the generalization of Transformer language models. We present a new method called self-distilled quantization (SDQ) that minimizes accumulative quantization errors and outperforms baselines. We apply SDQ to multilingual models XLM-R$_{\text{Base}}$ and InfoXLM$_{\text{Base}}$ and demonstrate that both models can be reduced from 32-bit floating point weights to 8-bit integer weights while maintaining a high level of performance on the XGLUE benchmark. Our results also highlight the challenges of quantizing multilingual models, which must generalize to languages they were not fine-tuned on.
\end{abstract}

\section{Introduction}
\begin{figure}
    \centering
    \includegraphics[scale=0.65]{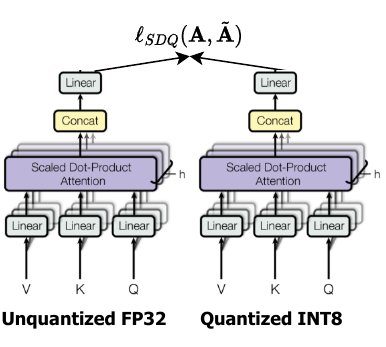}
    \vspace{-.5em}
    \caption{Self-Attention Self-Distilled Quantization}
    \label{fig:sdq_diagram}
    \vspace{-1.5em}
\end{figure}

\vspace{-0.5em}
A main aim of neural network quantization is to reduce the size and computational demands of a model while maintaining its performance. There are two main approaches: quantization-aware training (QAT)~\cite{banner2018scalable,chin2020one,faghri2020adaptive,kim2020position,wang2018training} and post-training quantization (PTQ)~\cite{neill2020overview,bondarenko2021understanding,kim2021bert,dettmers2022gpt3}. Both of these approaches have limitations in terms of dealing with accumulative quantization errors that are propogated within the layers of a neural network during the forward pass~\cite{zhao2019improving,fan2020training}. To address this issue, we propose a method called Self-Distilled Quantization (SDQ) that combines self-attention and output distillation with quantization to compress large language models. SDQ involves injecting quantization noise into the student network during training and distilling knowledge from a fine-tuned teacher network from both its final output and outputs of intermediate self-attention layers. By distilling knowledge of the self-attention layers, as depicted in~\autoref{fig:sdq_diagram}, we further reduce the compounding effect of quantization errors in the network. We use SDQ for self-attention models and demonstrate its effectiveness in compressing multilingual models XLM-R$_{\text{Base}}$ and InfoXLM$_{\text{Base}}$, achieving high compression rates while maintaining performance on the XGLUE benchmark. Lastly, we identify that quantization error is largest at the output of self-attention modules.
% SDQ involves injecting quantization noise into the student network during training while distilling knowledge from a fine-tuned teacher network from both its final output \emph{and} the outputs of each self-attention block, as depicted in~\autoref{fig:sdq_diagram}. 
%  However, most past work has focused on KD and quantization in the context of Convolutional Neural Networks in computer vision and more importantly, they only distill knowledge from the \emph{output} of the teacher. We argue that this is a limiting factor for very large networks and the accumulation of quantization errors can be reduced earlier in the network if a distillation loss is used on the intermediate layers between the student and teacher network.
% Unlike previous work that focuses on convolutional layers~\cite{mishra2017apprentice,polino2018model,kim2019qkd}, we focus on using SDQ for self-attention based models by reducing the quantization error where it is most apparent in the network. This is identified by a post-training analysis of where the quantization error is largest within the self-attention block across all layers. 
% If enough space add this back in 
% Moreover, since the student network has the same capacity as the teacher network before quantization, it enables the student to learn the distilled labels while applying quantization noise. In contrast, previous work argued that distilling knowledge into a smaller student network that is being quantized during fine-tuning induces too strong of a regularization effect~\cite{kim2019qkd}.
\iffalse
We find that SDQ sets state-of-the-art (SoTA) results and when used with current QAT methods it improves over the original QAT method without distillation. 
We focus on quantizing cross-lingual Transformer models, namely XLM-RoBERTa~\citep{conneau2019unsupervised} and InfoXLM~\cite{chi2020infoxlm}. To our knowledge, this is the first study that focuses on quantizing cross-lingual language models and in turn how quantization effects generalization on more than one language. We now move to a background on quantization for neural networks.
\fi

\vspace{-0.7em}
\section{Related Work} 
\vspace{-0.5em}
Combining quantization and distillation has been previously explored by~\citet{mishra2017apprentice}, who used three different schemes to combine low bit precision and knowledge distillation (KD) using a 4-bit ResNet network.~\citet{polino2018model} used a distillation loss with respect to a quantized teacher network to train a student network, and also proposed differentiable quantization, which optimizes the location of quantization points through SGD.~\citet{zhou2017incremental} used iterative quantization, supervised by a teacher network, to retrain an FP-32 model with low precision convolution weights (binary, ternary, and 4 bits).~\citet{kim2019qkd} used QAT and fine-tuning to mitigate the regularization effect of KD on quantized models. Q8BERT~\cite{zafrir2019q8bert} and fully Quantized Transformer~\cite{prato2019fully} applied QAT with the Straight-Through Estimator to approximate non-differentiable quantization in INT-8 format. We now describe the methodology of SDQ. %, while I-BERT (Kim et al., 2021)

% Fully Quantized Transformer (Prato et al., 2019) also applied QAT with STE for Transformers on Neural Machine Translation, achieving results similar to the original model. I-BERT (Kim et al., 2021) extended this approach by also approximating nonlinear operations in integer format for pure INT-8 inference.

% In the below related work we discuss state of the art (SoTA) QAT approaches, .

\vspace{-0.5em}
\section{Methodology}
\vspace{-0.5em}
We begin by defining a dataset $\mathcal{D}: = \{(X_i, y_i)\}^D_{i=1}$ with samples $s_i = (X_i, \vec{y}_i)$, where each $X_i := (\vec{x}_1, \ldots, \vec{x}_N)$ and $\vec{x}_i \in \mathbb{R}^{d}$ is the $i$-th vector. For structured prediction $y_i \in \{0, 1\}^{N \times d_y}$ and for single and pairwise sentence classification, $y_i \in \{0, 1\}^{d_y}$, where $d_y$ is the number of classes.
Let $\vec{y}^S=f_{\theta}(X_i)$ be the output prediction ($y^{S} \in \mathbb{R}^{d_y}$) from the student $f_{\theta}(\cdot)$ with pretrained parameters $\theta:= \{\mat{W}_l, \vec{b}_l\}_{l=1}^{L}$ for $L$ layers and the outputs of self-attention blocks are denoted as $\vec{A}_l$. 
The loss function for standard classification fine-tuning is defined as the cross-entropy loss $\ell_{CE}(\vec{y}^S, \vec{y})$.
%:= -\frac{1}{d_y}\sum_{i=1}^{d_y} \vec{y}_{d_y}\log(\vec{y}^s_{d_y})$.
% where for a single sample $s_i$, $\mathcal{L}: \mathcal{Y} \times \mathbb{R}^n \to \mathbb{R}$. 
% When $m$ is set to $1$, PQ is equivalent to vector quantization (VQ) and when $m$ is equal to $C_{in}$, it is the scalar k-means algorithm. 
%The main benefit of PQ is its expressivity: each column $w_j$ is mapped to a vector in the product $C = C \times \cdots \times C$, thus PQ generates an implicit codebook of size $k_m$.
\paragraph{Self-Distilled Quantization}
For self-distilled quantization, we also require a fine-tuned teacher network $f_{\Theta}$, that has been tuned from the pretrained state $f_{\theta}$, to retrieve the soft teacher labels $y^{T} := f_{\Theta}(\vec{x})$, where $y^{T} \in \mathbb{R}^C$ and $\sum_c^C y^{T}_c=1$. The soft label $\vec{y}^{T}$ can be more informative than the one-hot targets $\vec{y}$ used for standard classification as they implicitly approximate pairwise class similarities through logit probabilities.
% After fine-tuning $f_{\theta}$ through stochastic gradient descent (SGD), the output activations of $f_{\Theta}$ are used to perform self-distillation.
The Kullbeck-Leibler divergence (KLD) $\ell_{\mathrm{KLD}}$ is then used with the main task cross-entropy loss $\ell_{CE}$ to express $\ell_{\mathrm{SDQ}_{\text{KLD}}}$ as shown in \autoref{eq:sdq_loss}, 
\begin{equation}
\begin{split}\label{eq:kld_loss}
\ell_{\mathrm{SDQ}_{\text{KLD}}} \text{=}\, \ell_{\mathrm{CE}}(\vec{y}^{S}, \vec{y})\text{+}
\alpha \tau^2 D_{\mathrm{KLD}}\big(\vec{y}^{S}, \vec{y}^{T}\big)   
\end{split}    
\end{equation}

where $D_{\mathrm{KLD}}(\vec{y}^{S},\vec{y}^{T}) = \mathbb{H}( \vec{y}^T) - \vec{y}^{T} \log(\vec{y}^{S})$, $\mathbb{H}(\vec{y}^T)=\vec{y}^{T}\log(\vec{y}^{T})$ is the entropy of the teacher distribution and $\tau$ is the softmax temperature. Following~\cite{hinton2015distilling}, the weighted sum of the cross-entropy loss and the KLD loss $\ell_{\mathrm{SDQ}_{\text{KLD}}} \text{=}\ell_{\mathrm{CE}}(\vec{y}^{S}, \vec{y}) \text{+}\alpha \tau^2 D_{\mathrm{KLD}}\big(\vec{y}^{S}, \vec{y}^{T}\big)$ is used as our main SDQ-based KD loss baseline, where $\alpha \in [0, 1]$. However, $D_{\mathrm{KLD}}$ only distils the knowledge from the soft targets of the teacher but does not directly reduce accumulative quantization errors of the outputs of successive self-attention layers. This brings us to our proposed attention-based SDQ loss $\ell_{\mathrm{SDQ}_{\text{Att-KLD}}}$ shown in~\autoref{eq:sdq_loss},
\begin{equation}
\begin{split}\label{eq:sdq_loss}
\ell_{\mathrm{SDQ}_{\text{Att-KLD}}} \text{=} \ell_{\mathrm{CE}}(\vec{y}^{S}, \vec{y}) \text{+}
\alpha \tau^2 D_{\mathrm{KLD}}\big(\vec{y}^{S}, \vec{y}^{T}\big) \\ 
+\beta \frac{1}{LH}\sum_{l=1}^{L}\sum_{h=1}^{H}\ell_{\mathrm{Attention}}\big(\mat{A}^{S}_{lh},\mat{A}^{T}_{lh}\big)  
\end{split}    
\end{equation}
where $\alpha$ and $\beta$ are regularization terms and $\ell_{\text{Attention}}$ computes the loss between the student and teacher outputs of each self-attention block in $L$ layers and $H$ attention heads per layer. We also consider two baselines, $\ell_{\text{SDQ}_\text{Att}}$ which is the same as~\autoref{eq:sdq_loss} without $\alpha \tau^2 D_{\mathrm{KLD}}(\vec{y}^{S}, \vec{y}^{T})$ and $\ell_{\text{SDQ}_\text{Hid}}$ which applies the Mean Squared Error (MSE) loss between the hidden state outputs instead of the attention outputs. The gradient of $D_{\text{KLD}}(\cdot,\cdot)$ is expressed as $\frac{\partial D_{\text{KLD}}(\vec{y}^{S}_i, \vec{y}^{T}_i)}{\partial \vec{y}^S_i} = \tau(\vec{y}^S_i/\tau - \vec{y}^T_i/\tau)$ and as $\tau \to \infty$, the gradient is approximately $1/(d_y\vec{y}^S_i - \vec{y}^T_i)$. Similarly, the gradient of the MSE loss on a single self-attention output in layer $l$ and head $h$ is $1/n_{lh}(\vec{a}^S_j - \vec{a}^T_j)$ for a single sample input $x$. Hence, we see the connection between derivatives between the KLD loss and the MSE loss when combining them in a single objective. We now move to desribing how SDQ is used in two QAT methods. 

\iffalse
We can then expressed the gradient step as,

\begin{equation}\label{eq:grad}
\begin{split}
    \mat{W}_{(t+1)} := \mat{W}_t - \eta \mathbb{E}_{
(X,y)\sim \mathcal{D}}(\tau(\vec{y}^S/\tau - \vec{y}^T/\tau) \\
+ (\mat{a}^S-\mat{a}^T)) - \nabla_{\mat{W}_t}\vec{y}^S
\end{split}
\end{equation}
\fi

%
\iffalse
Note that \emph{scaled} dot-product is used (normalization by $\sqrt{d_k}$) to avoid vanishing gradients of the $\mathrm{Softmax}$, which may occur when $d_k$ is large. 

The parameters for the $j$-th attention head $\mat{K}^j, \mat{V}^j \in \mathbb{R}^{d \times l}$, $\mat{U}^j \in \mathbb{R}^o$ for $j = 1,\ldots, n_a$ where $n_a$ is the number of attention heads. Then we summarize the formulation of multi-headed self-attention as Equation \eqref{eq:multi_head_att},
%
\begin{equation}\label{eq:multi_head_att}
\begin{split}
\mat{Z}^j & = \text{Softmax}\Big(\frac{\mat{Q}\mat{K}^j}{\sqrt{d_k}}(\mat{V}^j)^{\top}\mat{Q}^{T}\Big)\mat{Q}\mat{U}^j \\
\tilde{\mat{Z}} & = \mathrm{Concat}(\mat{Z}^1, \ldots \mat{Z}^n_a) \\
\mat{Z} & = \mathrm{Feedforward}(\mathrm{LayerNorm}(\tilde{\mat{Z}} + \mat{Q}))    
\end{split}
\end{equation}
%
where $\mat{Z}^{j} \in \mathbb{R}^{n \times d_{a}}$ and $\tilde{\mat{Z}} \in \mathbb{R}^{n \times d_{a}n_{a}}$, with $d_{a}$ being the dimensionality of the self-attention output.
\fi

\paragraph{Iterative Product Distilled Quantization}
We first consider using SDQ with iPQ~\cite{stock2019and}. This is achieved by quantizing $m$ subvectors for each $k$ columns of $\mat{W}$ where a codebook for each $k$ subvectors is learned to map each subvector to its nearest neighbor in the learned codebook $C \in \mathbb{R}^{k\times d}$ where $k$ is the number of codewords. 
%Any column $\mat{W}_{[:,i]}$ is mapped to its quantized version $\phi(\mat{W}_{[:,i]}) = (C_{[j,1]} \ldots, C_{[j,m]})$ where $[i,1]$ denotes the index of the codeword assigned to the first subvector of $\mat{W}_{[:,j]}$. 
The codebook is updated by minimizing $||\mat{W} - \tilde{\mat{W}}||_2^2 = \sum_i^d || \mat{W}_{[:, i]} - \phi(\mat{w}_{[:, i]})||^2_2$ where $\phi(\cdot)$ is the quantization function. This objective can be efficiently minimized with the k-means algorithm and the codewords of each layers are updated with SGD by averaging the gradients of each assigned block of weights. This is done iteratively from the bottom layers to the top layers throughout training where the upper layers are finetuned while the lower layers are progressively being quantized~\cite{stock2019and}. When using iPQ with SDQ, omitting the KLD loss and cross-entropy loss, the objective is $\ell_{\text{SDQ}_{\text{iPQ}}} =\sum_{l=1}^{L-F}\Big[||\mat{W}_l - \tilde{\mat{W}}_l||^2_2 + \frac{\beta}{L\text{-}F}\sum_i^d(\mat{A}^{S}_{l,i} - \mat{A}^T_{l,i})^2\Big]$ where $F$ is the number of finetuned layers (non-quantized) at that point in training. Hence, SDQ progressively quantizes the layers throughout training when used with iPQ.

\paragraph{Block-Wise Distilled Quantization Noise}
For the majority of our QAT-based experiments we use Quant-Noise~\cite{fan2020training}. Quant-Noise is a SoTA QAT method that applies (fake) block-wise quantization noise at random to each weight matrix. Concretely, blocks of weights $b_{kl}$ in $\mat{W}_l$ are chosen at random at a rate $p$ and quantization noise is added to the chosen blocks. 
\begin{figure}[ht]
\begin{subfigure}{.245\textwidth}
  \centering
  \includegraphics[width=.975\linewidth]{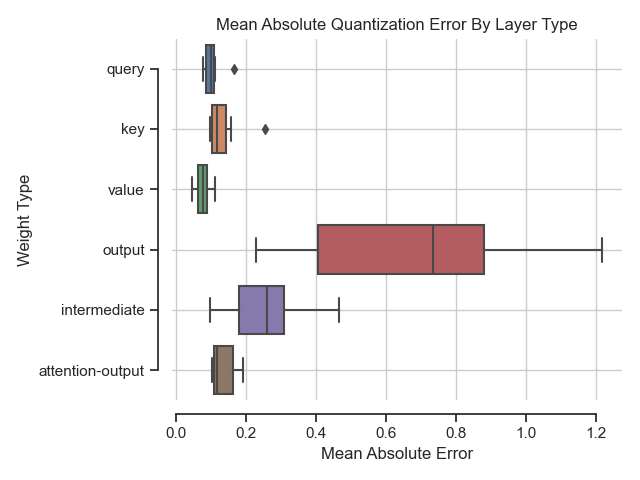}
  \caption{Quantization Error}
  \label{fig:q_error}
\end{subfigure}%
\begin{subfigure}{.245\textwidth}
  \centering
  \includegraphics[width=.975\linewidth]{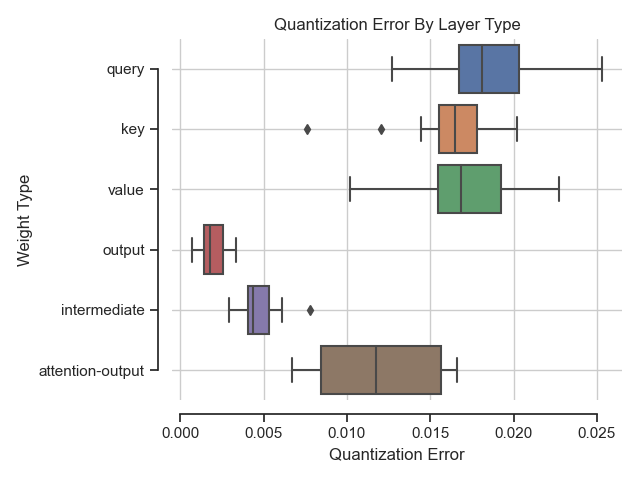}
  \caption{F-Norm Elementwise Errors}
  \label{fig:f_norm}
\end{subfigure}
\vspace*{-4mm}
\caption{Dynamic Quantization of InfoXLM$_{\mathrm{Base}}$ After Quantization Aware Fine-Tuning on XNLI.}
\label{fig:xnli_qat_by_layer}
%\vspace*{-2mm}
\end{figure}

\begin{table}[ht]
\centering
\resizebox{\linewidth}{!}{
\begin{tabular}{llll|c|c}
\toprule
\textbf{Student} & \textbf{Quant Method} & \textbf{Teacher} &  \textbf{Quant Method} & \textbf{en} & \textbf{Avg.} \\

\midrule
XLM-R$_{\text{Base}}$~\citeauthor{conneau2019unsupervised} & - & - & - & 84.6 & 74.5 \\
XLM-R$_{\text{Base}}$ & - & - & - & 83.9 & 73.9 \\
InfoXLM$_{\text{Base}}$ & - & - & - & 84.1 & 74.6 \\

InfoXLM$_{\text{Base}}$ & PTQ$_{\text{Dynamic}}$ & - & - & 81.7 & 71.4 \\
XLM-R$_{\text{Base}}$ & PTQ$_{\text{Dynamic}}$ & - & - & 80.1 & 72.5 \\

\midrule
\midrule
XLM-R$_{\text{Base}}$ & QNAT & - & - & 82.1 & 70.5 \\
InfoXLM$_{\text{Base}}$ & QNAT & - & - & 83.7 & 73.0 \\

\midrule
XLM-R$_{\text{Base}}$ & QNAT$_{\text{KLD}}$ & XLM-R$_{\text{Base}}$ & - & 83.4 & 72.5 \\
XLM-R$_{\text{Base}}$ & QNAT$_{\text{KLD}}$ & InfoXLM$_{\text{Base}}$ & - & 84.4 & 73.3 \\
InfoXLM$_{\text{Base}}$ & QNAT$_{\text{KLD}}$ & InfoXLM$_{\text{Base}}$ & - & 83.9 & 73.6 \\
InfoXLM$_{\text{Base}}$ & QNAT$_{\text{Att}}$ & InfoXLM$_{\text{Base}}$ & - & 84.1 & 73.2 \\
InfoXLM$_{\text{Base}}$ & QNAT$_{\text{Att-KLD}}$ & InfoXLM$_{\text{Base}}$ & - & 84.1 & \underline{\textbf{73.8}} \\

\midrule
InfoXLM$_{\text{Base}}$ & QNAT$_{\text{Att}}$ & InfoXLM$_{\text{Base}}$ & QNAT-PTQ & 83.3 & 72.1 \\
InfoXLM$_{\text{Base}}$ & QNAT$_{\text{Hid}}$ & InfoXLM$_{\text{Base}}$ & QNAT-PTQ & 81.1 & 70.7 \\
InfoXLM$_{\text{Base}}$ & QNAT$_{\text{KLD}}$ & InfoXLM$_{\text{Base}}$ & QNAT-PTQ & 83.7 & 73.1 \\
InfoXLM$_{\text{Base}}$ & QNAT$_{\text{Att-KLD}}$ & InfoXLM$_{\text{Base}}$ & QNAT-PTQ & 83.9 & 73.4 \\

\bottomrule
\multicolumn{6}{l}{The best performance obtained are marked in {\bf bold}.}
\end{tabular}
}
\caption{XNLI standard and zero-shot test accuracy using (Fake) Quantization-Aware Training with INT-8 Post-Training (Real) Dynamic Quantization.}
\label{tbl:xnli_detailed_results}
\vspace{-1.5em}
\end{table}

\vspace{-0.5em}
\section{Empirical Results}
\vspace{-0.5em}

\begin{figure}[ht]
\begin{subfigure}{.14\textwidth}
  \centering
  \includegraphics[width=.975\linewidth]{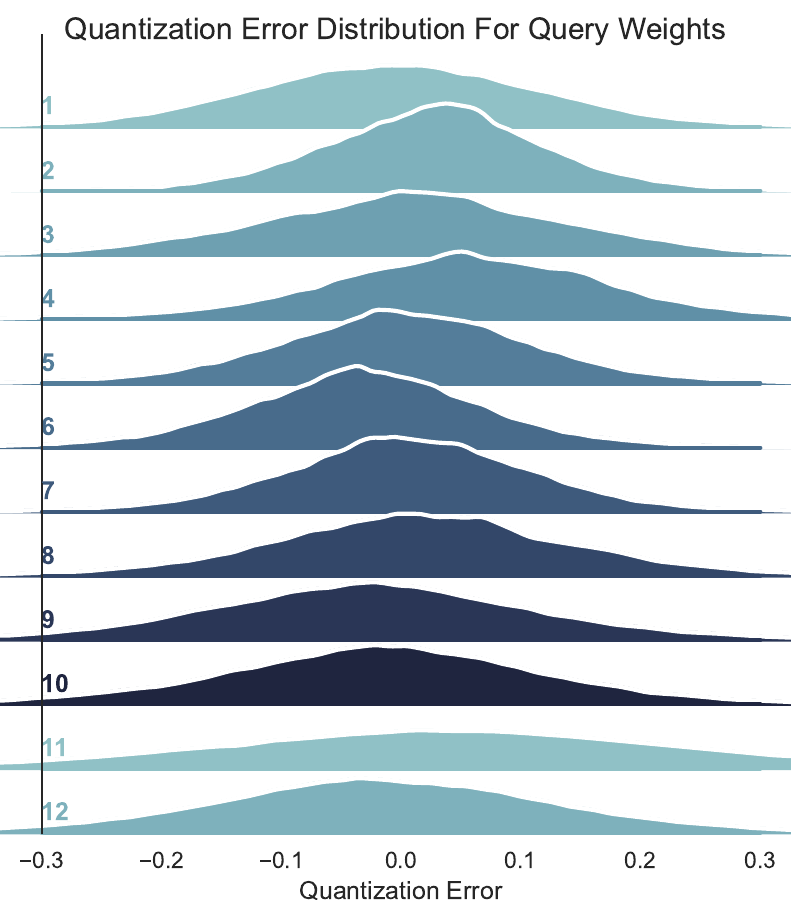}
  \caption{Query}
  \label{fig:query}
\end{subfigure}%
\begin{subfigure}{.17\textwidth}
  \centering
  \includegraphics[width=.975\linewidth]{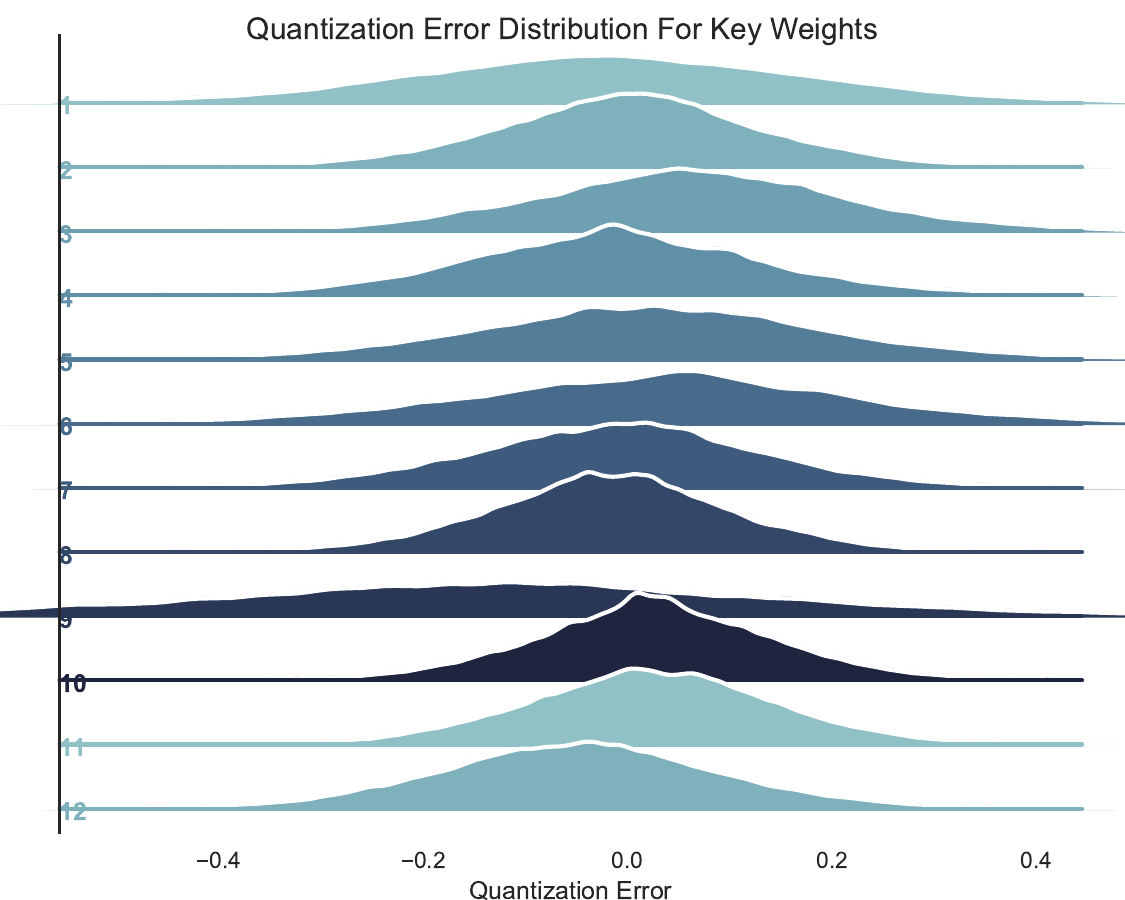}
  \caption{Key}
  \label{fig:key}
\end{subfigure}
\hfill
\begin{subfigure}{.15\textwidth}
  \centering
  \includegraphics[width=.975\linewidth]{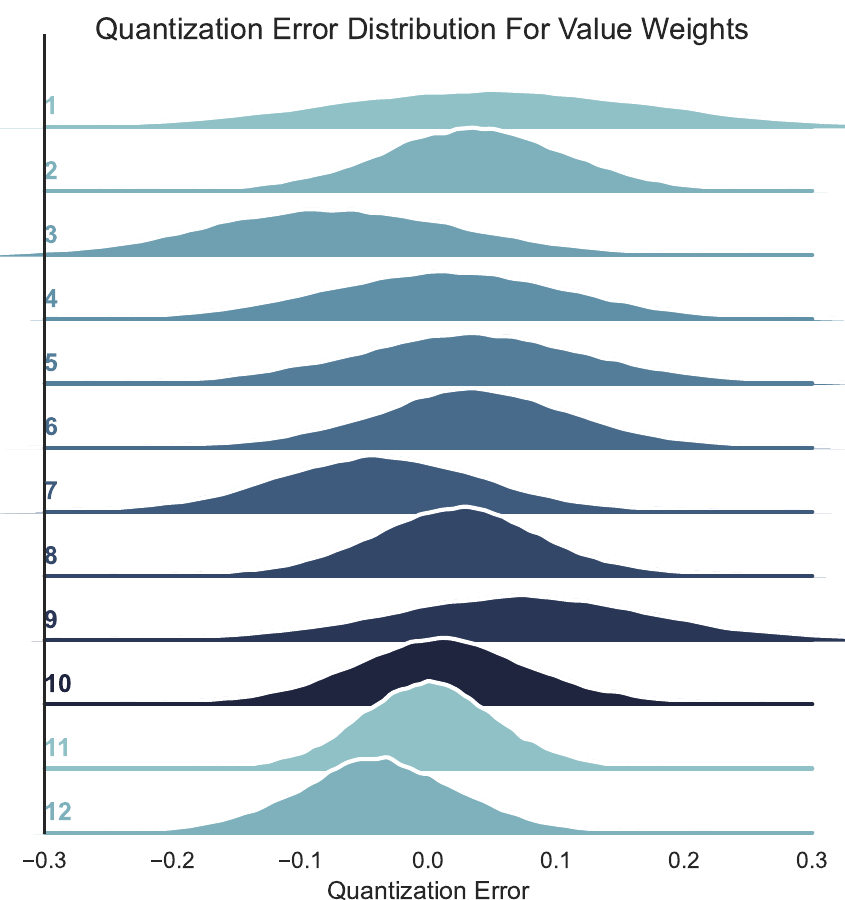}
  \caption{Value}
  \label{fig:value}
\end{subfigure}
\begin{subfigure}{.16\textwidth}
  \centering
  \includegraphics[width=.975\linewidth]{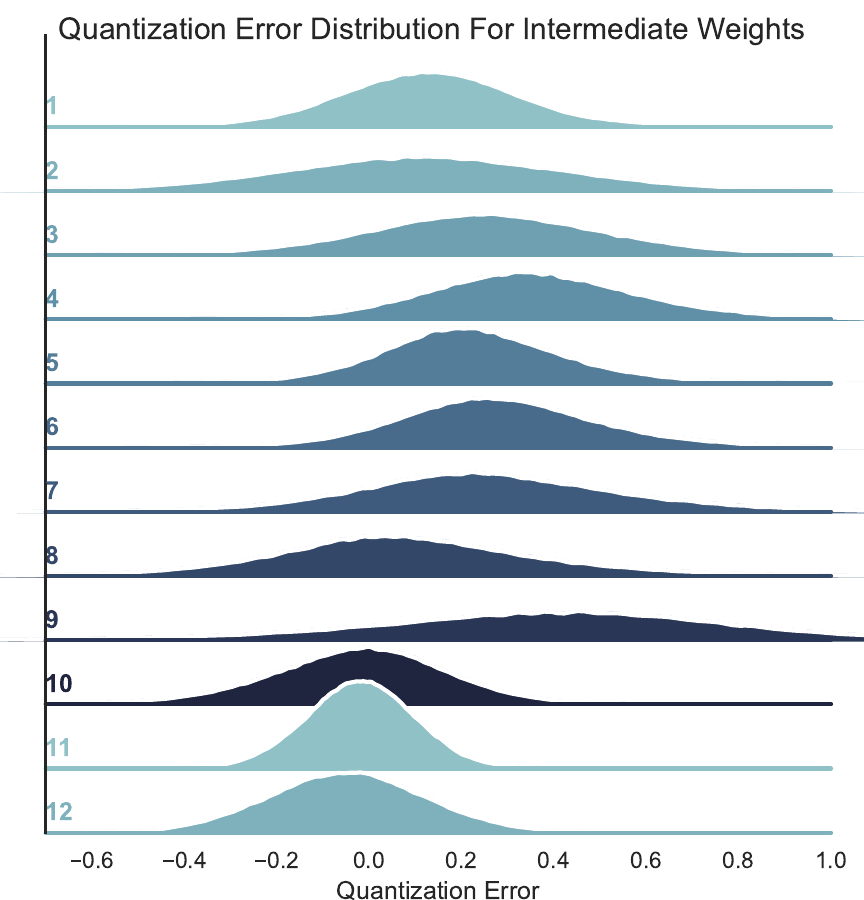}
  \caption{Intermediate Out}
  \label{fig:intermediate_output}
\end{subfigure}%
\begin{subfigure}{.16\textwidth}
  \centering
  \includegraphics[width=.975\linewidth]{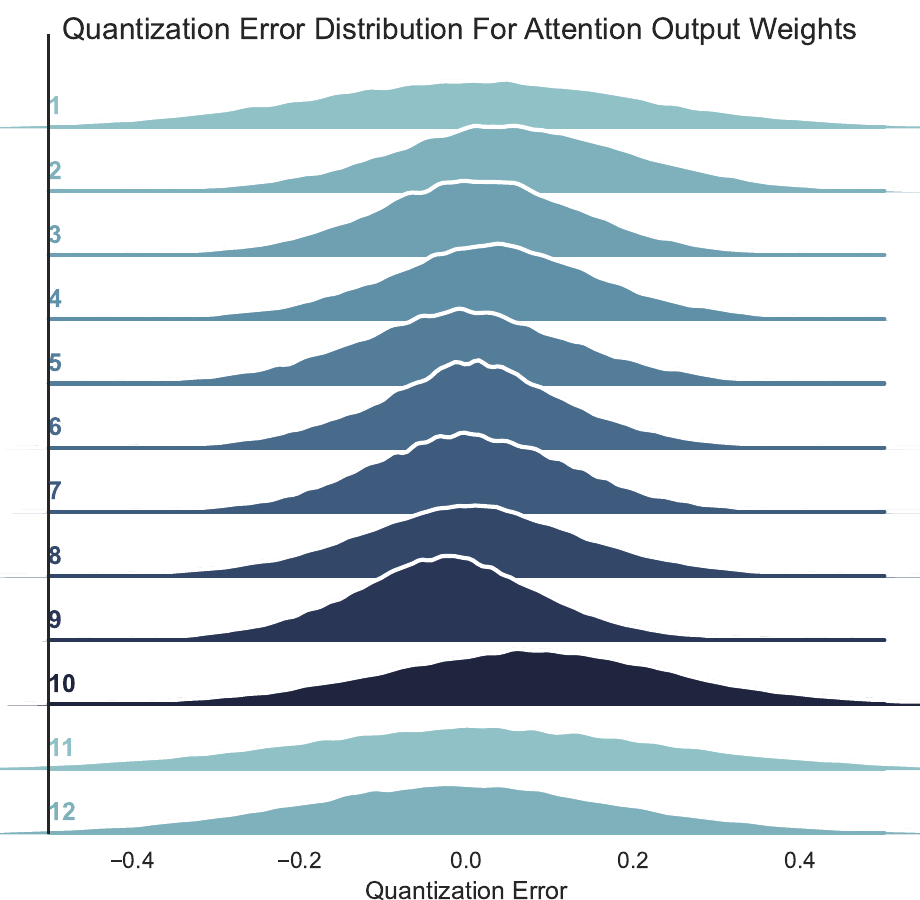}
  \caption{Attention Output}
  \label{fig:attention_output}
\end{subfigure}%
\begin{subfigure}{.17\textwidth}
  \centering
  \includegraphics[width=.9\linewidth]{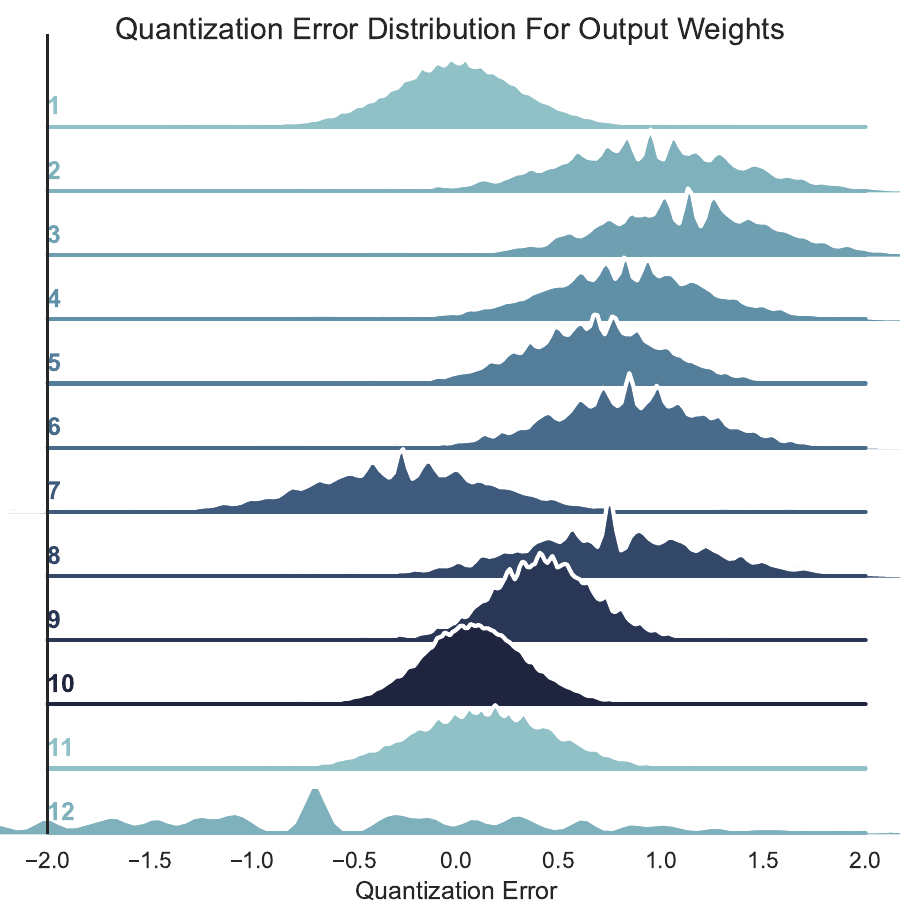}
  \caption{Output}
  \label{fig:output}
\end{subfigure}
\vspace*{-5mm}
\caption{Dynamic Quantization of InfoXLM$_{\mathrm{Base}}$ After Quantization Aware Fine-Tuning on XNLI.}
\label{fig:xnli_qat_by_ind}
\vspace*{-2mm}
\end{figure}

\iffalse
\begin{figure*}[ht]
\begin{subfigure}{.33\textwidth}
  \centering
  \includegraphics[width=.975\linewidth]{images/quant_by_layer_ind/xnli/query_p.pdf}
  \caption{Query}
  \label{fig:query}
\end{subfigure}%
\begin{subfigure}{.33\textwidth}
  \centering
  \includegraphics[width=.975\linewidth]{images/quant_by_layer_ind/xnli/key_p.pdf}
  \caption{Key}
  \label{fig:key}
\end{subfigure}
\hfill
\begin{subfigure}{.33\textwidth}
  \centering
  \includegraphics[width=.975\linewidth]{images/quant_by_layer_ind/xnli/value_p.pdf}
  \caption{Value}
  \label{fig:value}
\end{subfigure}
\begin{subfigure}{.33\textwidth}
  \centering
  \includegraphics[width=.975\linewidth]{images/quant_by_layer_ind/xnli/intermediate_output_p.pdf}
  \caption{Intermediate Output}
  \label{fig:intermediate_output}
\end{subfigure}%
\begin{subfigure}{.33\textwidth}
  \centering
  \includegraphics[width=.975\linewidth]{images/quant_by_layer_ind/xnli/attention_output_p.pdf}
  \caption{Attention Output}
  \label{fig:attention_output}
\end{subfigure}%
\begin{subfigure}{.33\textwidth}
  \centering
  \includegraphics[width=.8\linewidth]{images/quant_by_layer_ind/xnli/output_p.pdf}
  \caption{Output}
  \label{fig:output}
\end{subfigure}
\vspace*{-2mm}
\caption{\textbf{Dynamic Quantization of InfoXLM$_{\mathrm{Base}}$ After Quantization Aware Fine-Tuning on XNLI}.}
\label{fig:xnli_qat_by_ind}
\vspace*{-1mm}
\end{figure*}
\fi
\begin{table}[t]
\begin{center}
    % \scriptsize
    \resizebox{1.\linewidth}{!}{
    \begin{tabular}[b]{ll|l|cccccccc|l}
    \toprule[1.25pt]
    \textbf{Student} & \textbf{Teacher} & \textbf{Mem} & \textbf{XNLI} & \textbf{NC} & \textbf{NER} & \textbf{PAWSX} & \textbf{POS} & \textbf{QAM} & \textbf{QADSM} & \textbf{WPR} &  \textbf{Avg.} \\ 
    \midrule

\midrule

X & - & 1.22 & 73.9 & 83.2 & 83.8 & 89.3 & 79.7 & 68.4 & 68.3 & 73.6 & 77.5 \\
I & - & 1.22 & 74.6 & 83.6 & 85.9 & 89.6 & 79.8 & 68.6 & 68.9 & 73.8 & 78.1 \\
\midrule

\midrule
X-PTQ$_{\text{Dynamic}}$ & - & 0.52 & 71.4 & 81.5 & 82.9 & 87.1 & 76.1 & 66.3 & 65.8 & 68.2 & 74.9 \\

I-PTQ$_{\text{Dynamic}}$ & - & 0.52 & 72.5 & 81.8 & 83.0 & 87.8 & 75.8 & 66.6 & 66.1 & 68.7 & 75.3 \\
\midrule

X-QNAT & - & 0.52 & 70.5 & 81.8 & 83.0 & 87.4 & 78.4 & 66.8 & 66.9 & 70.4 & 75.7 \\
I-QNAT & - & 0.52 & 73.0 & 82.1 & 83.1 & 87.8 & 78.0 & 67.2 & 67.2 & 70.8 & 76.2 \\
\midrule
X-QNAT$_{\text{KLD}}$ & X & 0.52 & 72.5 & 82.0 & 83.2 & 88.1 & 78.8 & 67.1 & 67.2 & 70.7 & 75.8 \\
X-QNAT$_{\text{KLD}}$ & I & 0.52 & 73.3 & 82.1 & 82.8 & 88.2 & 78.3 & 67.3 & 67.5 & 70.5 & 75.9 \\
I-QNAT$_{\text{KLD}}$ & I & 0.52 & 73.6 & 82.6 & 83.1 & 88.4 & 79.5 & 67.6 & 67.9 & 71.8 & 76.8 \\
I-QNAT$_{\text{Att}}$ & I & 0.52 & 73.2 & 82.4 & 83.0 & 88.3 & 78.3 & 67.8 & 67.7 & 71.7 & 76.6 \\
I-QNAT$_{\text{Att-KLD}}$ & I & 0.52 & \textbf{73.8} & \textbf{82.8} & \textbf{83.4} & 88.8 & 79.5 & \textbf{67.9} & 68.0 & 72.4 & $\underline{\textbf{77.1}}$ \\

\midrule
I-QNAT$_{\text{Att}}$ & I$_{\text{QNAT-PTQ}}$ & 0.52 & 72.1 & 82.1 & 83.1 & 89.2 & 78.8 & 68.0 & 67.8 & 71.9 & 76.6 \\
I-QNAT$_{\text{Hid}}$ & I$_{\text{QNAT-PTQ}}$ & 0.52 & 70.7 & 81.9 & 82.4 & 88.8 & 78.4 & 67.3 & 68.0 & 71.4 & 76.1 \\
I-QNAT$_{\text{KLD}}$ & I$_{\text{QNAT-PTQ}}$ & 0.52 & 73.1 & 82.3 & 83.0 & 88.4 & 79.2 & 67.6 & 67.9 & 72.1 & 76.7 \\
I-QNAT$_{\text{Att-KLD}}$ & I$_{\text{QNAT-PTQ}}$ & 0.52 & 73.4 & 82.5 & 83.3 & \textbf{88.9} & \textbf{79.6} & \textbf{67.9} & \textbf{68.2} & \textbf{72.6} & $\underline{\mathbf{77.1}}$ \\
\bottomrule[1.25pt]
\end{tabular}
}
\caption{XGLUE INT-8 Zero-Shot Quantization Results with Post-Training Dynamic Quantization.}
\label{tab:quant_all}
\vspace{-2em}
\end{center}
\end{table}

We begin by referring the reader to the supplementary material for the experimental setup in~\autoref{sec:a2} and ~\autoref{sec:a3}.
% train_runtime': 8393.007, 'train_samples_per_second': 119.147
Before discussing the main results on XGLUE, we first analyse the mean absolute quantization error and the Frobenius norm of the elementwise difference in self-attention blocks between an INT-8 dynamically quantized InfoXLM$_{\text{Base}}$ and an unquantized FP-32 InfoXLM$_{\text{Base}}$ in~\autoref{fig:xnli_qat_by_layer}. We see in~\autoref{fig:q_error} that the output layer contains the largest mean absolute error across each layer and highest error variance. In contrast, query, key, value (QKV) parameters have much smaller error. However, since most of the parameters are found in the QKV layers, the sum of the quantization error is larger, as seen in~\autoref{fig:f_norm}. This motivates us to focus on the output of the self-attention block when minimizing quantization errors with our proposed loss in~\autoref{eq:sdq_loss} as the mean error is higher near the output as it accumulates errors from previous layers in the block.
This is also reflected in the parameter distribution of each layer type across all layers in
\autoref{fig:xnli_qat_by_ind}, where the x-axis is the mean absolute quantization error and the y-axis is the layer indices. We see the quantization noise is more apparent on the output layer as the Gaussian distrbutions are non-smooth and have clear jitter effect. % We also note that the lower layers (2-6) of the quantization output parameters have higher error than upper layers. % With the observations, we now move to the first set of results on XNLI.

\paragraph{XNLI Per Language Results}
~\autoref{tbl:xnli_detailed_results} shows the baselines and our SDQ methods applied to XLM-R$_\text{Base}$ and InfoXLM$_\text{Base}$. Here, both models are only trained on the English language and hence the remaining languages in the evaluation set test the zero-shot performance after INT8 quantization (apart from the first 3 rows that show FP-32 fine-tuned results). On average, we find that best student networks results are found when distilling using QNAT$_{\text{Att-KLD}}$ SDQ with the outputs of an FP-32 teacher for InfoXLM$_\text{Base}$ at 73.8\% test accuracy points, where the original FP-32 InfoXLM$_{\text{Base}}$ achieves 74.6\%. Additionally we see that QNAT$_{\text{Att-KLD}}$ improves over QNAT$_{\text{KLD}}$ distillation, indicating that attention output distillation improves the generalization of the INT-8 student model. We also found that largest performance drops correspond to languages that less pretraining data and morphologically rich (Swahili, Urdu, Arabic), while performance in English for the best INT-8 XLM-R$_{\text{Base}}$ (84.4\%) is within 0.2\% of the original network (84.6\%) and the best InfoXLM$_{\text{Base}}$ that uses QNAT$_{\text{Att-KLD}}$ is on par with the FP-32 results.

\iffalse
We also find that languages which have less training data for pretraining and are morphologically rich (Swahili, Urdu and Arabic) suffer the largest drops in performance. From this we posit that FP-32 resolution allows for capacity that maintains information about more complex sentences. 
In contrast, the standard supervised learning evaluation on English achieves comparable results to the original FP-32 results across all baselines and our proposed SDQ (the best INT-8 XLM-R$_{\text{Base}}$ (84.4\%) is within 0.2\% of the original network (84.6\%) and the best InfoXLM$_{\text{Base}}$ that uses QNAT$_{\text{Att-KLD}}$ is on par with the FP-32 results).
\fi

 \begin{figure*}
     \centering
     \includegraphics[width=0.925\textwidth]{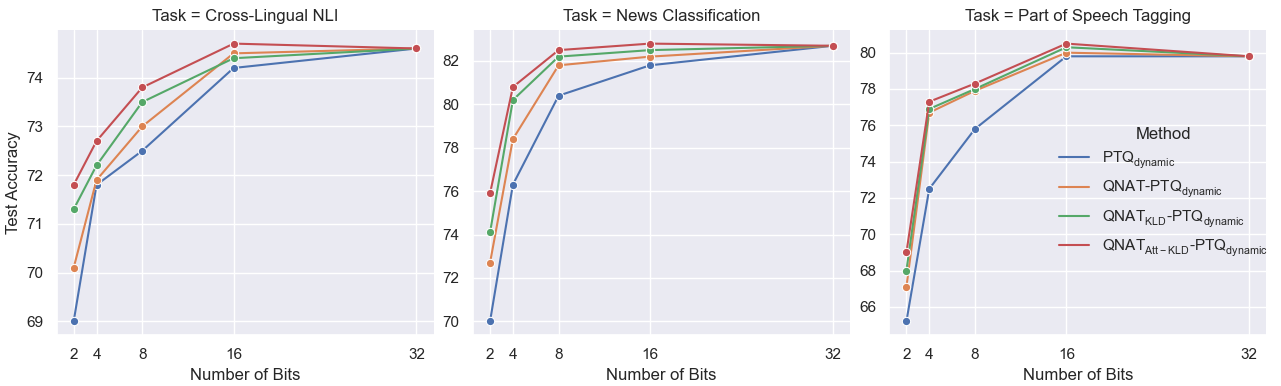}
    \vspace{-.5em}
     \caption{Accuracy versus Number of Bits for XGLUE Tasks with FP-32-16 and INT-8-4-2 formats.}
     \label{fig:acc_v_nbits}
     \vspace{-1.em}
 \end{figure*}

\paragraph{Quantization Results on {\em XGLUE}.}

We show the per task test performance and the {\em understanding score} (i.e average score) on XGLUE for quantization baselines and our proposed SDQ approaches in Table \ref{tab:quant_all} (for brevity we denote InfoXLM$_{\text{Base}}$ as I and XLM-R$_{\text{Base}}$). Our proposed QNAT$_\text{Att-KLD}$ achieves the best average (Avg.) score and per task performance for all tasks, using a fine-tuned InfoXLM$_{\text{Base}}$ (XNLI, NC, NER and QAM) and a fine-tuned InfoXLM$_{\text{Base}}$ trained with QuantNoise and dynamically quantized post-training (PAWSX, POS, QAM, QADSM and WPR). We also find that QNAT$_{\text{Att-KLD}}$ improves over QNAT$_{\text{KLD}}$, highlighting that the attention loss is improving quantized model performance. 

\paragraph{Performance versus Compression Rate}
~\autoref{fig:acc_v_nbits} shows how the performance changes for four approaches, including two of our proposed objectives (QNAT$_{\text{KLD}}$ and QNAT$_{\text{Att-KLD}}$), when training InfoXLM$_\text{Base}$. As before, PTQ$_{\text{dynamic}}$ is a dynamically quantization fine-tuned InfoXLM$_{\text{Base}}$ and QNAT-PTQ$_{\text{dynamic}}$ is the same as PTQ$_{\text{dynamic}}$ except fine-tuned also using QuantNoise.
Unlike our previous results, here we apply fake quantization at inference to achieve compression lower than INT-8 and be comparable to previous work~\cite{fan2019reducing}. We see that performance is generally well maintained up until 8 bits. However, performance significantly degrades for all quantization methods for 4 and 2 bit weights. We find that QNAT$_{\text{Att-KLD}}$ maintains higher performance when compared to the baselines and directly quantizing with no QAT (PTQ$_{\text{dynamic}}$) leads to the poorest results, also reflected in ~\autoref{tab:quant_all} results with real dynamic quantization at inference time.
% Two strong predictors of how much test performance drops as a percentage of the original FP-32 model is the amount of training data and the number of languages used for evaluation. For example, POS performance is the largest of the three tasks (has least amount of training data and the 17 languages used for evaluation), XNLI is second (the most training data but 15 languages for evaluation) and NC has the lowest performance drop (moderate training data and only 5 languages for evaluation). 
\begin{table}[t]
\begin{center}
    % \scriptsize
    \resizebox{1.\linewidth}{!}{
    \begin{tabular}[b]{ll|cccc|l}
    \toprule[1.25pt]
    \textbf{Student} & \textbf{Teacher} & \textbf{XNLI} & \textbf{NC} & \textbf{NER} & \textbf{POS} & \textbf{Avg.} \\ 
    
\midrule
\midrule
- & - & 74.6 & 83.6 & 85.9 & 79.7 & 81.0 \\
\midrule
iPQ$_{\text{Scalar}}$ & - & 69.1 & 79.4 & 81.9 & 76.3 & 76.7 \\
iPQ$_{\text{Scalar-KLD}}$ & Standard & 70.4 & 80.1 & 82.3 & 76.9 & 77.4 \\
iPQ$_{\text{Scalar-KLD}}$ & iPQ$_{\text{Scalar}}$ & 70.8 & 80.7 & 82.6 & 79.4 & 78.4 \\
iPQ$_{\text{Scalar-Att-KLD}}$ & Standard & 72.2 & 80.4 & 82.5 & 77.4 & 78.1 \\
iPQ$_{\text{Scalar-Att-KLD}}$ & iPQ$_{\text{Scalar}}$ & 71.3 & 80.4 & 82.9 & 79.6 & \underline{\textbf{78.6}} \\
\midrule
iPQ$_{\text{EM}}$ & - & 69.1 & 79.4 & 81.9 & 76.3 & 76.7 \\
iPQ$_{\text{EM-KLD}}$ & Standard & 70.4 & 80.1 & 82.3 & 76.9 & 77.4 \\
iPQ$_{\text{EM-KLD}}$ & iPQ$_{\text{EM}}$ & 72.8 & 81.6 & 82.8 & 79.8 & 79.3 \\
iPQ$_{\text{EM-Att-KLD}}$ & Standard & 73.2 & 82.3 & 82.7 & 79.1 & 79.3 \\
iPQ$_{\text{EM-Att-KLD}}$ & iPQ$_{\text{EM}}$ & 73.1 & 82.5 & 83.0 & 79.2 & \underline{\textbf{79.5}} \\

\midrule
QNAT & - & 70.5 & 81.8 & 83.3 & 78.4 & 78.5 \\
QNAT$_{\text{KLD}}$ & Standard & 73.2 & 82.6 & 83.1 & 79.5 & 79.6 \\
QNAT$_{\text{KLD}}$ & QNAT & 73.1 & 82.3 & 83.0 & 79.2 & 79.4  \\
QNAT$_{\text{Att-KLD}}$ & Standard & 73.8 & 82.8 & 83.4 & 79.5 & \underline{\textbf{79.9}} \\
QNAT$_{\text{Att-KLD}}$ & QNAT & 73.4 & 82.5 & 83.3 & 79.6 & 79.7 \\
\bottomrule[1.25pt]
\end{tabular}
}
\caption{SoTA INT-8 Iterative Product Quantization methods with and without SDQ for InfoXLM$_{\text{Base}}$.}
\label{tab:sota_with_sdq}
\vspace{-1.5em}
\end{center}
\end{table}
\paragraph{Ablation with Current QAT Methods}
~\autoref{tab:sota_with_sdq} shows the results from XGLUE tasks where the first two columns describe how the student and teacher networks are trained and ``Standard'' refers to standard FP-32 fine-tuning. This includes iPQ~\cite{stock2019and} with scalar quantization (iPQ$_\text{Scalar}$), iPQ that uses expectation maximization to create the codebook during training (iPQ$_\text{EM}$) and previous results of QuantNoise (QNAT) as a reference point. In this setup, we only apply the attention loss, $\ell_{\text{Attention}}$, to the layers that are quantized during iPQ. 
% We note that this baseline is also first application of iPQ to Transformers
In all cases, adding that SDQ distillation of the classification output and the self-attention outputs improves the average performance. 

\iffalse
\subsection{Summary of Results.} From our experiments we find that minimizing the MSE between the output of quantized self-attention modules and unquantized outputs (i.e distillation) leads to consistently better performance of quantized models. Moreover, most of the quantization errors after dynamic quantization are largely attributed to the query, key and value (QKV) parameters, but this is only due to the discrepancy between the number of parameters used for these layers with remaining fully-connected output layers within the self-attention block. We also find quantization error per layer type is largest on the output linear layer with a significantly larger mean error and variance when compared to QKV outputs. This is due to the accumulation of quantization errors within the self-attention block itself. Hence, this motivated us to focus on the output of the attention block to apply the distillation loss, to achieve a desired trade-off between the total number of losses applied during training (only 1 per self-attention block) and its direct effect on generalization performance of quantized models. 
\fi

\vspace{-0.5em}
\section{Conclusion}
\vspace{-0.5em}
In this paper we proposed an attention-based distillation that minimizes accumulative quantization errors in fine-tuned masked language models. We identified that most of the quantization errors accumulate at the output of self-attention blocks and the parameter distribution of the output layer is effected more by quantization noise. 
The proposed distillation loss outperforms baseline distillation without the attention loss and the resulting INT-8 models are within 1 understanding score points on the XGLUE benchmark with \emph{real} quantization post-training. Moreover, fine-tuning the teacher network with quantization-aware training can further improve student network performance on some of the tasks.
Further compression can be achieved up to 4-bit and 2-bit weights but performance steeply degrades as the network capacity is drastically reduced coupled with the models having to generalize to multiple languages it was not trained on. %In future work, we aim to further improve quantized model performance in the context of the zero-shot cross-lingual setting.

\vspace{-0.5em}
\section{Limitations}
\vspace{-0.5em}
\paragraph{Dataset and Experimental Limitations.}
The datasets and tasks we focus on are from the XGLUE benchmark~\cite{liang2020xglue}. The structured prediction tasks, namely Named Entity Recognition (NER) and Part of Speech (PoS) Tagging, both have a limited number of training samples at 15k and 25.4k samples respectively. This is due to the difficulty in annotating on the token level, however it can still be viewed as a limitation when compared to the remaining sentence-level tasks the majority of tasks have at least 100k samples. 

\paragraph{Methodological Limitations.}
Below are a list of the main methodological limitations we perceive of our work:
\begin{itemize}
    \item Our method requires a teacher model that is already trained on the downstream task which can then be used to perform knowledge distillation. This is limiting when there are constraints on the computing resources required to produce the quantized model. 
    \item We have focused on the problem of reducing accumulative qunatization errors which become more apparent the deeper a network is. However, this problem is intuitvely lessened when the model is shallow (e.g 3-4 layers) but perhaps wider. Hence the results may be less significant if the model is shallower than what we have experimented in this work.
    \item By introducing the distillation loss we require an additional regualrization term $\beta$ to be optimally set, relative to the main distillation loss $\alpha$. This can be viewed as a potential limitation has it introduced an additional hyperparameter to be searched to obtain best results on a given task.  
    \item Lastly, since intermediate layer outputs of the teacher network are required for self-attention distillation, we have to perform two forward passes during training. Since standard KLD distillation only requires the output logits, it is common to store the training data teacher logits, eliminating the need to perform two forward passes at training data. However, this is not an option with self-atttention outputs as the storage required offline scales with the number of self-attention heads, number of layers and the size of the training data. 
\end{itemize}

\section{Ethics Statement}
Here we briefly discuss some ethical concerns of using such compressed models in the real world, specifically the two techniques used in this work, quantization and knowledge distillation.~\citet{hooker2020characterising} have found that compressed models can amplify existing algorithmic bias and perform very poorly on a subset of samples while the average out-of-sample accuracy is maintained close to the uncompressed model. This general finding for pruning and quantization may be also extrapolated to our work (including distillation), hence it is important to recognize that our work, much like the remaining literature on compression, may have ethical concerns with regards to algorithmic bias and how that effects downstream tasks. However, smaller models are more cost-efficient and thus become more widely available to the general public. To summarize, it is important to analyse any aforementioned bias amplification for subsets of samples for downstream tasks compressed models are used for. 

% Entries for the entire Anthology, followed by custom entries
\bibliography{anthology,acl2023}
\bibliographystyle{acl_natbib}

\appendix

\section{Supplementary Material}
\label{sec:appendix}
\subsection{Self-Attention in Transformers}

Consider a dataset $D = \{(X_i, y_i)\}_{i=1}^{m}$ for $D \in \mathcal{D}$ and a sample $s:= (X, y)$ where the sentence $X:= (x_1, \ldots x_n)$ with $n$ being the number of words $x \in X$. 
We can represent a word as an input embedding $\vec{x}_w \in \mathbb{R}^d$, which has a corresponding target vector $\vec{y}$. 
In the pre-trained transformer models we use, $X_i$ is represented by 3 types of embeddings; word embeddings ($\mat{X}_w \in \mathbb{R}^{n \times d}$), segment embeddings ($\mat{X}_s \in \mathbb{R}^{n \times d}$) and position embeddings ($\mat{X}_p \in \mathbb{R}^{n \times d}$), where $d$ is the dimensionality of each embedding matrix. 
The self-attention block in a transformer mainly consists of three sets of parameters: the query parameters $\mat{Q} \in \mathbb{R}^{d \times l}$, the key parameters $\mat{K} \in \mathbb{R}^{d \times l}$ and the value parameters $\mat{V} \in \mathbb{R}^{d \times o}$. For 12 attention heads (as in XLM-R$_{\text{Base}}$ and InfoXLM$_{\text{Base}}$), we express the forward pass as follows:

\begin{gather}\label{eq:sa_block}
    \overrightarrow{\mat{X}} = \mat{X}_w + \mat{X}_s + \mat{X}_p  \\
    \overrightarrow{\mat{Z}} := \bigoplus_{i=1}^{12}   \text{softmax}\big(\overrightarrow{\mat{X}} \mat{Q}_{(i)} \mat{K}^{T}_{{(i)}} \overrightarrow{\mat{X}}^{T}\big)\overrightarrow{\mat{X}}\mat{V}_{(i)}  \\
    \overrightarrow{\mathbb{Z}} = \text{Feedforward}(\text{LayerNorm}(\overrightarrow{\mat{Z}} + \overrightarrow{\mat{X}})) \\
        \overleftarrow{\mathbb{Z}} = \text{Feedforward}(\text{LayerNorm}(\overleftarrow{\mat{Z}} + \overleftarrow{\mat{X}})) 
\end{gather}

The last hidden representations of both directions are then concatenated $\mathbb{Z}' :=  \overleftarrow{\mathbb{Z}}  \bigoplus \overrightarrow{\mathbb{Z}'}$ and projected using a final linear layer $\mat{W} \in \mathbb{R}^{d}$ followed by a sigmoid function $\sigma(\cdot)$ to produce a probability estimate $\hat{y}$, as shown in \eqref{eq:cf_layer}. Words from (step-3) that are used for filtering the sentences are masked using a \texttt{[PAD]} token to ensure the model does not simply learn to correctly classify some samples based on the association of these tokens with counterfacts. 
% WordPiece embeddings~\cite{wu2016google} are used and a special classification token \texttt{[CLS]} is used to signify the end of each sentence. 
A linear layer is then fine-tuned on top of the hidden state, $\vec{h}_{X, \texttt{[CLS]}}$ emitted corresponding to the \texttt{[CLS]} token. This fine-tunable linear layer is then used to predict whether the sentence is counterfactual or not, as shown in \autoref{eq:cf_layer}, where $\cB \subset D$ is a mini-batch and $\mathcal{L}_{ce}$ is the cross-entropy loss.

\begin{align}\label{eq:cf_layer}
 \mathcal{L}_{ce} := \frac{1}{|\cB|} \sum_{(X,y) \in \cB} \vec{y} \log \big( \sigma (\vec{h}_{X,\texttt{[CLS]}} \cdot \mat{W}) \big)
\end{align}

\textbf{Configurations} We use XLM-R$_{\text{Base}}$ and InfoXLM$_{\text{Base}}$, which uses 12 Transformer blocks, 12 self-attention heads with a hidden size of 768. 
The default size of 512 is used for the sentence length and the sentence representation is taken as the final hidden state of the first [CLS] token.
% This model is already pre-trained and we fine-tune a linear layer $\mat{W}$ on top of BERT, which is fed to through a sigmoid function $\sigma$ as $p(c|h) = \sigma(\mat{W}\vec{h})$ where $c$ is the binary class label and we maximize the log-probability of correctly predicting the ground truth label.

\subsection{Experimental Setup and Hardware Details}\label{sec:a2}
Below describes the experimental details, including model, hyperparameter and quantization details.
We choose modestly sized cross-lingual language models as the basis of our experiments, namely XLM-R$_{\text{Base}}$~\cite{conneau2019unsupervised} and InfoXLM$_{\text{Base}}$~\cite{chi2020infoxlm}, both approximately 1.1GB in memory and these pretrained models are retrieved from the \href{https://huggingface.co/models}{huggingface model hub}.

We choose both XLM-R$_{\text{Base}}$ and InfoXLM$_{\text{Base}}$ because they are relatively small Transformers and are required to generalized to languages other than the language used for fine-tuning. Hence, we begin from a point that model are already relatively difficult to compress and are further motivated by the findings that larger overparameterized networks suffer less from PTQ to 8-bit integer format and lower~\citep{jacob2018quantization,krishnamoorthi2018quantizing}. 

For both XLM-R$_{\text{Base}}$ and InfoXLM$_{\text{Base}}$ the hyper-parameters are set as follows: 768 hidden units, 12 heads, GELU activation, a dropout rate of 0.1, 512 max input length, 12 layers in encoder. The Adam Optimizer with a linear warm-up~\cite{vaswani2017attention} and set the learning rate to 2e-5 for most tasks. For all sentence classification tasks the batch size is set to 32 and we fine-tune with 10 epochs. For POS Tagging and NER, we fine-tune with 20 epochs and set the learning rate to 2e-5. We select the model with the best average results on the development sets of all languages. For SDQ-based models, we report the best performing model for $\alpha \in [0.1, 0.2, 0.5, 0.8]$ and $\beta \in [10, 100, 200, 500]$. All experiments are carried out on Tesla V100-SXM2 32 Gigabyte GPUs~\cite{nvidia2017nvidia} with no constraint on GPU hours used on these machines. In all reported results, we report the best (max) result from 8-16 different runs when searching for $\alpha$ and $\beta$ depending on each particular task.

\iffalse
\section{Intended Use of Existing Artifacts}
The artifacts we use in this work are the datasets from the XGLUE benchmark that are used for evaluation and the pretrained models themselves.
\fi

\subsection{Model Configuration and Hyperparameter Settings}\label{sec:a3}
XLM-R$_{\text{Base}}$ and InfoXLM$_{\text{Base}}$ uses 12 Transformer blocks, 12 self-attention heads with a hidden size of 768.  The default size of 512 is used for the sentence length and the sentence representation is taken as the final hidden state of the first [CLS] token. A fine-tuned linear layer $\mat{W}$ is used on top of both models, which is fed to through a softmax function $\sigma$ as $p(c|h) = \sigma(\mat{W}\vec{h})$ where $c$ is used to calibrate the class probability estimate and we maximize the log-probability of correctly predicting the ground truth label.

\autoref{tab:pretrain_hyperparam_transformer} shows the pretrained model configurations that were already predefined before our experiments. The number of (Num.) hidden groups  here are the number of groups for the hidden layers where parameters in the same group are shared. The intermediate size is the dimensionality of the feed-forward layers of the the Transformer encoder. The `Max Position Embeddings' is the maximum sequence length that the model can deal with.

\begin{table}[ht]
	\centering
	\resizebox{.5\textwidth}{!}{%
	\begin{tabular}{l|c|c}
		\toprule[1.pt]
		 Hyperparameters & XLM-R$_{\text{Base}}$ & InfoXLM$_{\text{Base}}$ \\
		\hline
		Vocab Size & 250002 &  250002\\
        Max Pos. Embeddings & 514 & 514\\
		Hidden Size & 3072 & 3072 \\
		Encoder Size & 768 & 768 \\
        Num. Hidden Layers & 12 & 12\\
        Num. Hidden Groups & 1 & 1 \\
        Num. Attention Heads & 12 & 12 \\

        Hidden Activations & GeLU & GeLU \\
        Layer Norm. Epsilon & $ 10^{-12}$ & $ 10^{-12}$\\
        Fully-Connected Dropout Prob. & 0.1 & 0.1\\

        Attention Dropout Prob. & 0 & 0\\
	\bottomrule[1.pt]
	\end{tabular}%
 }
	\caption{Model Hyperparameter Settings}
	\label{tab:pretrain_hyperparam_transformer}
\end{table}

We now detail the hyperparameter settings for transformer models and the baselines. We note that all hyperparameter settings were performed using a manual search over development data. 

\subsubsection{Transformer Model Hyperparameters}
We did not change the original hyperparameter settings that were used for the original pretraining of each transformer model. The hyperparameter settings for these pretrained models can be found in the class arguments python documentation in each configuration python file in the \url{https://github.com/huggingface/transformers/blob/master/src/transformers/} e.g configuration\_.py.% and are also summarized in \autoref{tab:pretrain_hyperparam_transformer}.
For fine-tuning transformer models, we manually tested different combinations of a subset of hyperparameters including the learning rates $\{50^{-4}, 10^{-5}, 50^{-5}\}$, batch sizes $\{16, 32, 128\}$, warmup proportion $\{ 0, 0.1\}$ and $\epsilon$ which is a hyperparameter in the adaptive momentum (adam) optimizer. 
Please refer to the huggingface documentation at \url{https://github.com/huggingface/transformers} for further details on each specific model e.g at \url{https://github.com/huggingface/transformers/blob/master/src/transformers/modeling_roberta.py}, and also for the details of the architecture that is used for sentence classification and token classification.
% Fine-tuning all language models with a sentence classifier took less than two and half hours for all models. 
\end{document}